\documentclass[conference]{IEEEtran}
\makeatletter
\def\ps@headings{%
\def\@oddhead{\mbox{}\scriptsize\rightmark \hfil \thepage}%
\def\@evenhead{\scriptsize\thepage \hfil \leftmark\mbox{}}%
\def\@oddfoot{}%
\def\@evenfoot{}}
\makeatother
\usepackage[utf8]{inputenc}
\usepackage{amsmath,amssymb,amsfonts}
\usepackage{algorithmic}
\usepackage{graphicx}
\usepackage{textcomp}
\usepackage{lscape}
\usepackage{booktabs}
\usepackage{multirow}
\usepackage{multicol}
\usepackage{longtable}
\usepackage{pifont}
\usepackage{subfigure}
\usepackage{graphicx}
\usepackage{comment}
\usepackage{balance}
\usepackage{amsmath}
\usepackage{color}
\usepackage{url}
\usepackage[linesnumbered,ruled,vlined]{algorithm2e}
\SetKw{KwBy}{by}
\SetKw{KwTo}{in}
\usepackage{mathtools}

\usepackage[table,xcdraw]{xcolor}
\usepackage{ulem}
\usepackage{cancel}
\usepackage[hyperindex,breaklinks]{hyperref}
\usepackage{subfigure}
\usepackage{breqn}
\usepackage{lipsum}
\usepackage{xurl}

\newcommand*{\affmark}[1][*]{\textsuperscript{#1}}

\begin{document}

\title{Harris Hawks Feature Selection in Distributed Machine Learning for Secure IoT Environments}

\author{\IEEEauthorblockN{Neveen Hijazi\affmark[1], Moayad Aloqaily\affmark[1], Bassem Ouni\affmark[2], Fakhri Karray\affmark[1], Merouane Debbah\affmark[2]} \\

%\affmark[1]Concordia University, QC, Canada. \\
\IEEEauthorblockA{
\affmark[1]Mohamed Bin Zayed University of Artificial Intelligence (MBZUAI), UAE \\
\affmark[2]Technology Innovation Institute (TII), Abu Dhabi, UAE\\
E-mails: \protect \affmark[1]\{neveen.hijazi; moayad.aloqaily; fakhri.karray\}@mbzuai.ac.ae}
\affmark[2]\{bassem.ouni; merouane.debbah\}@tii.ae \\

%\protect \textit{E-mails}: \{h.elayan; maloqaily\}@xanalytics.ca, mguizani@ieee.org, haythem@email.arizona.edu
\vspace{-1.0cm}

%\footnote{This work was jointly supported by Qatar University - Grant Number IRCC [2020-003]. The findings achieved herein are solely the responsibility of the authors.}
}

\maketitle

\begin{abstract}
The development of the Internet of Things (IoT) has dramatically expanded our daily lives, playing a pivotal role in the enablement of smart cities, healthcare, and buildings. Emerging technologies, such as IoT, seek to improve the quality of service in cognitive cities. Although IoT applications are helpful in smart building applications, they present a real risk as the large number of interconnected devices in those buildings, using heterogeneous networks, increases the number of potential IoT attacks. IoT applications can collect and transfer sensitive data. Therefore, it is necessary to develop new methods to detect hacked IoT devices. This paper proposes a Feature Selection (FS) model based on Harris Hawks Optimization (HHO) and Random Weight Network (RWN) to detect IoT botnet attacks launched from compromised IoT devices. 
Distributed Machine Learning (DML) aims to train models locally on edge devices without sharing data to a central server. Therefore, we apply the proposed approach using centralized and distributed ML models. Both learning models are evaluated under two benchmark datasets for IoT botnet attacks and compared with other well-known classification techniques using different evaluation indicators. The experimental results show an improvement in terms of accuracy, precision, recall, and F-measure in most cases. The proposed method achieves an average F-measure up to 99.9\%. The results show that the DML model achieves competitive performance against centralized ML while maintaining the data locally.
\end{abstract}
\begin{IEEEkeywords}
IoT, Harris Hawks Optimization, Cognitive Cities, Smart Buildings, Distributed Machine Learning.
\end{IEEEkeywords}

\section{Introduction}\label{sec:intro}
Internet of Things (IoT) is one of the most significant revolutions of the 21st century and can be simply described as a set of connected smart edge devices. These devices can be found in many applications we encounter across our daily lives including healthcare, smart cities, transportation systems, and smart grids \cite{%al2021intelligent,
aloqaily2019intrusion}.
IoT environments typically consist of collections of distributed heterogeneous devices with low computational memory, and they have the ability to expand with other interconnected networks, which results in many security concerns \cite{ghimire2022recent}.
The enormous increase in IoT devices and their lack of security guarantees has motivated attackers to launch attacks by setting up large-scale IoT botnet attacks.
A botnet is a network of devices connected to IoT devices that are infected by malware. In addition, compromised devices have a simple infection process, which makes every vulnerable device a bot candidate. Some of the most popular examples of IoT botnet attacks are Mirai, BASHLITE, and Phishing \cite{kolias2017ddos}. Consequently, this highlights the importance of increasing security measures to identify the most critical cyberattacks in a smart IoT environment.
%, as a result, achieving a balance between privacy and efficiency.% \cite{al2022overview}.

%%%%%%%%%%%%%%%%%%%%%%%%%%%%%

Intrusion Detection Systems (IDSs) are security systems that detect attacks and malicious activities in networks, such as botnet attacks. ML techniques can play an essential role in IDS structures to classify and predict attacks. Different ML techniques are used to detect anomalies and attacks in cybersecurity, such as K-Nearest Neighbor (KNN),
Support Vector Machines (SVM), and Decision Trees (DT) \cite{alzubi2022efficient,faris2019intelligent}.
However, IoT devices continuously produce vast amounts of data that is challenging to deal with using traditional ML techniques and requires complex models to process such as deep learning technique \cite{aloqaily2019intrusion}.
Therefore, there is a need for more efficient searching and learning algorithms, which has led to the emergence of metaheuristic algorithms.

Metaheuristics algorithms are a well-regarded choice to solve high-dimensional problem through the utilization of the feature selection technique \cite{chkirbene2020tidcs}. The objective of this technique is to generate a highly accurate general model while minimizing the number of selected features in order to reduce computational time. Therefore, various works have utilized metaheuristic algorithms in feature selection methods such as Genetic Algorithm (GA), Particle Swarm Optimization (PSO), and Harris Hawks Optimization (HHO) \cite{faris2019intelligent,hijazi2021parallel}. 

Distributed ML is a valuable approach, it aims to train one or more ML models for a network of users, each having a local dataset. Thus, distributed ML reduces data transmissions within centralized ML and minimizes the risks of privacy leaks. In addition, it can reduce the risk of a single point of failure compared to centralized ML  \cite{jiang2022anonymous}. Distributed ML can be applied in various applications such as healthcare \cite{zerka2021privacy} and wireless network \cite{ghimire2022recent}.
In this study, we propose a hybrid model based on the HHO and Random Weight Network (RWN) algorithms to detect botnet attacks launched from compromised IoT devices. The proposed approach performs wrapper FS based on two main stages: centralized and distributed ML. Wrapper FS consists of three repeated steps including search algorithm, learning algorithm, and an evaluation measure. In our work, we used HHO as a search algorithm to select the near-optimal feature subset of the input features and optimized the RWN structure simultaneously to increase the prediction power. Then, RWN is applied as a learning algorithm to fit the training dataset according to the selected solution. Finally, evaluation measures are applied to assess the quality of the selected solution. The proposed approach is compared with other metaheuristics approaches \cite{habib2020modified}.

The contributions of this work can be summarized as
\begin{enumerate}
    \item We propose an intrusion detection system based on HHO with RWN.
    \item We apply RWN as the base classifier in wrapper FS, unlike most of the literature that uses KNN.
    \item We utilize the HHO algorithm to achieve two different objectives simultaneously, including, near-optimal subset of features and number of hidden neurons in RWN.
    \item We leverage distributed ML to maintain data privacy and apply the proposed approach using a distributed ML scheme.

\end{enumerate}
The rest of the paper is organized as follows. Section \ref{sec:Preliminaries} presents the background and related works of Harris Hawks Optimization, Random Weight Network algorithms, and Decentralized machine learning. The methodology and proposed approach are described in detail in Section \ref{sec:methodology}. The conducted
experiments and the obtained results are discussed
in Section \ref{sec:results}. Finally, we conclude our work and provide suggestions for future directions in Section \ref{sec:conclusion}.

\section{Background and Related Work}\label{sec:Preliminaries}
\subsection{Harris Hawks Optimization (HHO)}
\label{sec:preHHO}
HHO is a meta-heuristic optimization algorithm
developed by A. Heidari et al. \cite{heidari2019harris}. The HHO algorithm was inspired by the behavior of Harris Hawks.
Precisely, HHO uses a multi-agent system where each agent represents a solution and collaborates with other agents to search for the global optimal solution. The algorithm starts with a set of initial solutions, and then updates the solutions by exchanging information and searching in a new direction. The algorithm terminates when a stopping criterion is met, such as reaching a maximum number of iterations or finding a solution that meets a certain fitness threshold. HHO uses a combination of global and local search techniques to balance the exploration and exploitation of a search space in order to find the global optimum solution. In the exploration phase, candidate solutions are randomly generated and evaluated for potential optimization. The selection of the solution generation strategy is based on a probability of either considering the location of already discovered solutions or relying on randomly determined positions.
The exploitation phase of the HHO algorithm is a key component that influences the optimization performance. In this phase, the algorithm uses the surprise pounce strategy to search for optimal solutions in the problem space. The algorithm follows four different chasing strategies: Soft Besiege, Hard Besiege, Soft Besiege with progressive rapid dives, and Hard Besiege with advanced fast dives. The choice of strategy is based on the energy of the prey, where the prey energy loses its energy while escaping the hawk. Thus, the hawks alternate between different exploitative behaviors based on the prey's energy during running.

HHO is applied in different and broad applications in the literature. For instance, O. Alzubi et al. \cite{alzubi2022efficient} proposed an SVM combined with HHO for Android malware detection. Their research relied on two targets, the first goal was to determine the best hyperparameters for the SVM, and the second was to determine the weights of the features to determine the most important ones, thus improving the detection process by using the HHO algorithm. Their approach was compared against existing techniques using four different datasets. The results showed superior performance over other metaheuristic algorithms and state-of-art classifiers.
Another work also mentioned the importance of HHO \cite{sokkalingam2022intelligent}. In this study, S. Sokkalingam et al. proposed three hybrid ML approaches with feature selection methods for IDS: SVM-PSO, SVM-HHO-PSO, and SVM-HHO. The performance of these methods was evaluated using the NSL-KDD dataset and compared with various classifiers. The experimental outcomes indicated that SVM-HHO-PSO achieved the best results compared to other methods.

\subsection{Random Weight Networks (RWN)}
\label{sec:preRWN}

RWN is a computational neural network model developed by Schmidt et al. \cite{schmidt1992feed} to handle the training process of a Single Hidden Layer Feedforward Neural Network (SLFN), where a node can be a subnet that consist of extra hidden node. RWN has overcome the problems of back-propagation, such as slow convergence and high probability of being trapped in a local minimum. In RWN, the input weights and the hidden biases are initialized randomly; then, the output weights are analytically computed using the Moore–Penrose generalized inverse method. 
Compared to other gradient descent methods used in SLFN, the RWN learning rate is speedy and has better generalization performance. Moreover, RWN does not need human intervention to set the parameters manually such as learning rate, number of epochs, etc. RWN has only one parameter that needs to be initialized, which is the number of hidden neurons. 

RWN is a powerful learning algorithm, which is used in a wide range of applications. For example, H. Faris et al. \cite{faris2019intelligent} proposed a hybrid email spam detection system based on GA and RWN. The proposed approach performs a wrapper FS method using an RWN classifier. The GA is used to select the optimal features and determine the number of neurons in the hidden layer of the RWN. Their approach detected spam emails and achieved excellent results in terms of accuracy, recall, and precision.
Another hybrid method proposed by E. Rawashdeh et al. in \cite{rawashdeh2021cooperative}. In this study, the authors depended on RWN and PSO. The aim of the work was to optimize the structure of the RWN and to find the best subset of features. PSO was used to find the best subset of features, as well as the number of hidden neurons in RWN. The proposed approach was evaluated on thirty datasets, compared with other state-of-the-art methods, and showed superior results in most datasets.

\subsection{Distributed Machine Learning (DML)}
\label{sec:preDML}
In this approach of learning, a generic model is distributed by the server to all clients or devices. The devices then customize the model through local training and testing with their own data, enabling predictions and insights  from live data generated by the device. This approach keeps the data local, reducing security and privacy concerns,  but the downside is that IoT devices are not well-suited for intensive computation \cite{ghimire2022recent}.

Distributed ML has gained more attention in recent years. For instance, F. Zerka et al. \cite{ zerka2021privacy} applied several ML algorithms using a distributed framework to preserve medical data privacy. Their approach was compared with the centralized ML approach using four small medical datasets. The results show that the distributed approach delivers similar performance to the centralized approach, while preserving privacy of the patient data. A. Tuladhar et al. \cite{tuladhar2020building} investigated a distributed ML model by ensemble using three classifiers: ANNs, SVM, and Random Forests (RF). Their approach was evaluated on four medical datasets. They found improvement in results when increasing the number of locally trained models in the ensemble. The authors suggested using the proposed approach on small datasets.

According to the No Free Lunch theorem (NLF) \cite{wolpert1997no}, there is no single optimization algorithm that outperforms the others in all optimization problems. Therefore, this field of research offers opportunities for improvement and the development of new techniques that can tackle a wider range of problems and applications. This will ultimately lead to more competitive and improved results when compared to current algorithms.
Overall, it can be understood from the literature that applying evolutionary algorithms with RWN is gaining more attention from researchers. HHO is a recent optimization algorithm that has been applied in different applications and has achieved promising results compared to other metaheuristic techniques. Moreover, it provides good exploration and exploitation. Therefore, it would be wise to use these properties with a robust learning algorithm, such as RWN. In conclusion, the proposed approach differs from previous studies by proposing a wrapper FS based on HHO and RWN algorithms to solve the IoT botnet detection problem, and applying the proposed approach within distributed ML that addresses IoT security and privacy issues.

\begin{figure}[]
\centering
\includegraphics[ width=0.50\textwidth ]{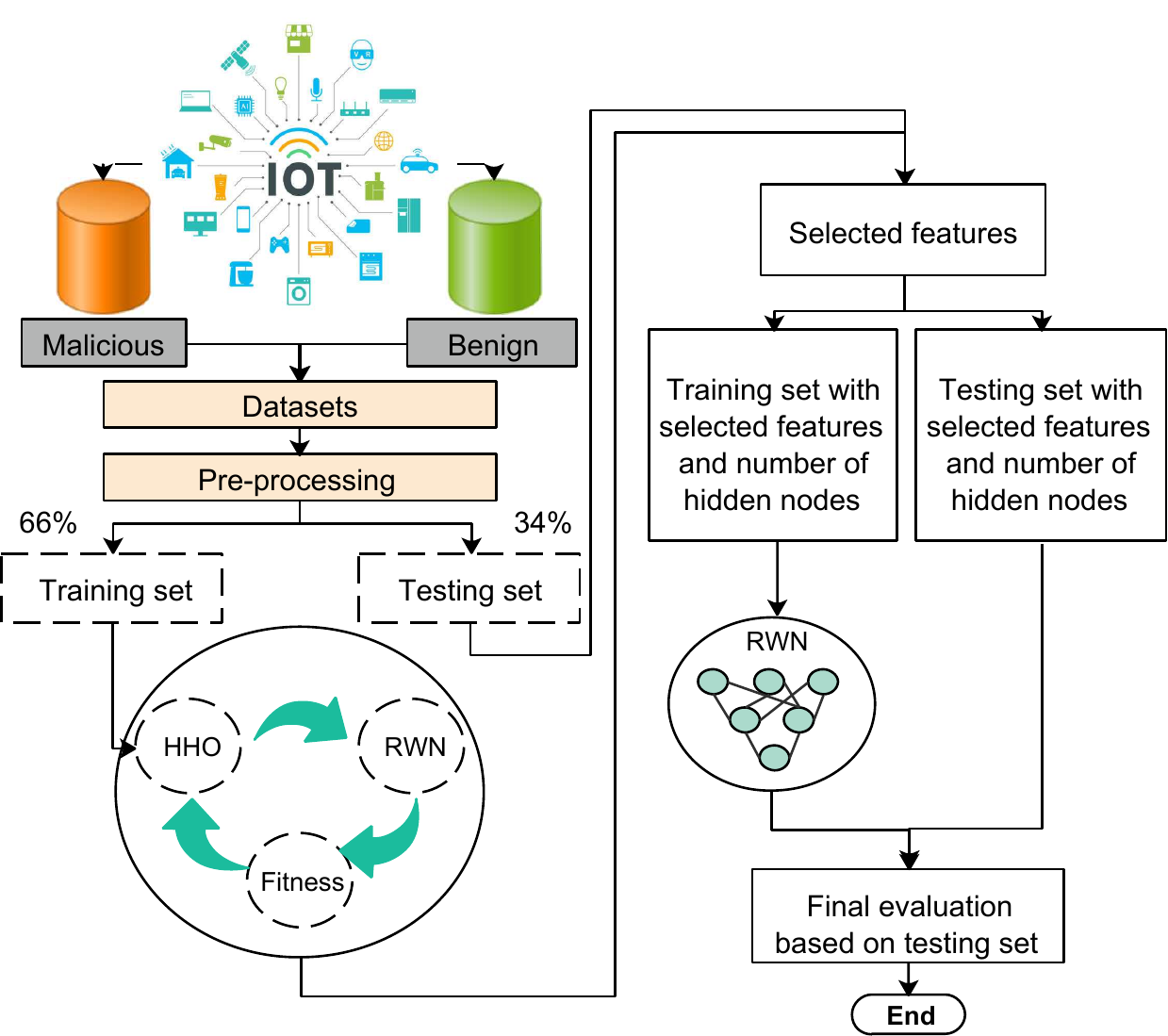}
\caption{An overview of the proposed approach.}
\label{fig:Method}       % Give a unique label
\end{figure}

\section{Methodology}
\label{sec:methodology}
%%%%%%%%%%%%%%%%%%%%%%%%
The proposed work is conducted based on a two-stage: centralized and distributed ML models. We apply the FS method, model construction, evaluation, and assessment in each stage. The main contribution lies in FS and the phases of model construction, where a new model based on HHO and RWN is proposed.
%%%%%%%%%%%%%%%%%%%%%%%%%
This section presents the details of the proposed approach, which is based on using HHO for FS while simultaneously optimizing the RWN structure.

\subsection{Data Preprocessing}
Two benchmark datasets were used for this work: N-BaIoT dataset \cite{meidan2018n} and Phishing legitimate dataset \cite{tan2018phishing}. The N-BaIoT dataset is obtained from the University of California at Irvine (UCI) machine learning repository. This dataset represents a real IoT network consisting of nine devices: a baby monitor, two doorbells, four security cameras, a thermostat, and a webcam. The second dataset was collected from Kaggle datasets. The dataset comprises 48 features extracted from 10,000 web pages, with 5000 being phishing pages and the other 5000 being legitimate ones \cite{tan2018phishing}.
Using the preliminary dataset without any preprocessing may impact the overall performance of the obtained model. Because we deal with massive and real datasets, they may contain inconsistent and redundant data. In order to get the best possible results, we apply data integration and data normalization. In the integration step, we integrated the data for the N-BaIoT dataset according to each device type. For example, the four different types of security cameras are merged into one file. All the devices in the N-BaIoT dataset have 115 features and 7,062,606 instances. These devices typically connect to IoT networks through Wi-Fi. For each device, data were collected during normal operation and various attacks conducted by the BASHLITE and Mirai botnet. The total number of attacks carried out by BASHLITE and Mirai were 555,932 and 2,838,272 respectively.
The integration step's benefit is reducing the number of generated models \cite{meidan2018n}.
For the normalization step, we apply the Min-Max normalization method to both datasets to make all the input values within a consistent scale [0,1]. 

\subsection{Wrapper Feature Selection based on HHO-RWN}
In this work, we apply an FS method on the training dataset to eliminate irrelevant features, reduce data dimensionality, and increase prediction performance.
Two well-known methods in feature selection are filter-based and wrapper-based approaches. In our work, we used a wrapper-based approach. Wrapper FS consists of three main components: the search algorithm, learning algorithm, and evaluation of the features. Wrappers are more potent than filters when accuracy is more important than speed. This is because they explore the relevant subset of features in the search space of a particular classifier, where most studies use the same classifier used in the learning process \cite{faris2019intelligent}. On the other hand, the drawback of the wrappers is that they require a high computational time.  Most studies use fast classifiers such as KNN as a primary classifier  \cite{faris2019intelligent}. 

In our proposed approach, we utilize RWN as the base classifier; in addition to the fast-learning property, RWN has good generalization performance. Our wrapper-based approach selects HHO to search for the relevant feature subsets. 
%HHO is utilized in different applications in the literature, such as FS, Intrusion detection, and so on, and showed promising performance on these applications.
In order to apply a metaheuristic optimization algorithm for a given task, two crucial design issues have to be addressed: the solution representation and the fitness function used to evaluate the solution, which are discussed below:

\begin{enumerate}
\item Solution Representation: A hawk in HHO represents a candidate solution for the targeted problem. A solution in our work consists of two parts. The first part is a set of binary bits 1 or 0, which indicate if the corresponding features are selected or not. The second part of the candidate solution consists of a group of binary bits that determine the number of hidden nodes in RWN. The length of the solution is L + M, where L is the number of features in the dataset, and M is the number of bits reserved to represent the maximum number of hidden neurons.
\item Fitness Function: A fitness function is needed to assess the quality of the solutions. In this work, we use a fitness function that addresses three objectives: increasing the classification accuracy, reducing the dimensionality of the dataset by selecting the relevant features, and decreasing the complexity of the RWN in terms of the number of hidden nodes. This fitness function is defined in Eq. \ref{equ:equ6}.
\begin{equation}
\begin{aligned}
Fitness &= \alpha Err + \beta \frac{f}{F} + \gamma \frac{n}{N} 
\label{equ:equ6}       % Give a unique label
\end{aligned}
\end{equation}
\end{enumerate}

The parameters $\alpha, \beta,$ and $\gamma$ are three factors within the range [0–1] that control the weight of each of the classification rate, number of selected features, and hidden neurons, respectively. $Err$ represents the classification error rate of the RWN classifier, $f$ denotes the number of selected features, $F$ is the total number of original features in the dataset, $n$ is the number of neurons determined by the evaluated HHO solution, and $N$ is the maximum number of possible neurons in the RWN. The maximum number of neurons is set to 1024, which is represented by ten elements in the solution.
Usually, the error rate is measured as $(Err = 1-Acc)$, where $Acc$ represents the classification accuracy. To consider the imbalanced datasets, we use the F-measure instead of accuracy in the error rate measure. Thus, the error rate in our case becomes $(Err = 1- F\_measure)$. After we obtained the selected features from the FS process, we must remove the irrelevant features and keep only the selected features from the training and testing data which will highly improve the performance measure.
The proposed approach is illustrated in Fig.\ref{fig:Method}. The
code is available in the GitHub \cite{Neveen}.

As mentioned earlier, the proposed approach was applied in two stages: centralized and distributed ML. In the centralized ML model, we trained the proposed approach and other traditional ML techniques by using the entire training dataset. Then we use the centralized ML model as a reference to compare the performance of each corresponding distributed ML model in each dataset.
In contrast, the distributed ML models, representing various IoT devices, were trained on local datasets. We split the training datasets by sampling without replacement method to ensure the diversity (i.e. different parts of the training datasets) for each user. So, each user has local data and applies the proposed approach and other ML techniques, thus distributed ML contributes to preserve the data privacy.

\section{Results and Discussion}\label{sec:results}
This section presents a performance evaluation of the proposed approach through experiments on the datasets. All the experiments were performed using a PC with Intel Core (TM) i7-165G7 2.8 GHz and 16 GB RAM. The proposed approach was implemented and tested in Python 3.9.12. Initially, the datasets are divided into ($66\%$ and $34\%$) for training and testing, respectively. The training dataset is further distributed into ($75\%$ and $25\%$) for training and validation, respectively.

Since Evolutionary Algorithms (EAs) are based on randomness, experiments must be repeated several times to reduce the random effect in EAs performance. Therefore, the proposed approach was tested under 30 independent runs, while the number of iterations was 100 and the population size was 200. Commonly, the population size should be large enough to guarantee the diversity of the solutions \cite{fukumoto2018study}. Most recent references have set the value of the population size to 200 or less \cite{hijazi2021parallel, habib2020modified}. Therefore, we used the maximum population size value of many studies. 
Also, many previous works have set the values of $\alpha, \beta,$ and $\gamma$, fitness function parameters to 0.99, 0.01, and 0.01, respectively \cite{faris2019intelligent, hijazi2021parallel}. Therefore, this paper uses the same values of these parameters in the fitness function.
In addition, we compare the performance with other known classifiers commonly used in the literature, including KNN, SVM, Adaboost, and DT. This experiment aims to show the motivation for using RWN as a primary classifier within the FS method and as a final classifier. 
Because the datasets are imbalanced, considering the accuracy ratio for evaluation is misleading. Therefore, we should examine other metrics such as recall, precision, and F-measure.
For all results, the best results were indicated by a bold typeface. The results of accuracy, recall, precision, and F-measure values of the centralized and distributed ML models are shown in Tables \ref{tab:comp} and \ref{tab:comp2}, respectively.
According to the HHO-RWN results in Table \ref{tab:comp}, we can see that all average accuracy values are enhanced compared to SVM and Adaboost classifiers. The accuracy rate has improved up to $2.6\%$ compared to HHO-SVM in Security camera device. 
Regarding the average precision results, we can observe that all values in HHO-RWN are higher than other classifiers. The precision enhancement rate reached $6.6\%$ in Phishing legitimate dataset. As for the average F-measure, HHO-RWN outperforms the other classifiers with 4 out of 5 devices in the N-BaIoT dataset. It also achieved the best average F-measure result in Phishing legitimate dataset.
\begin{figure}[t]
\centering
\includegraphics[width=0.50\textwidth]{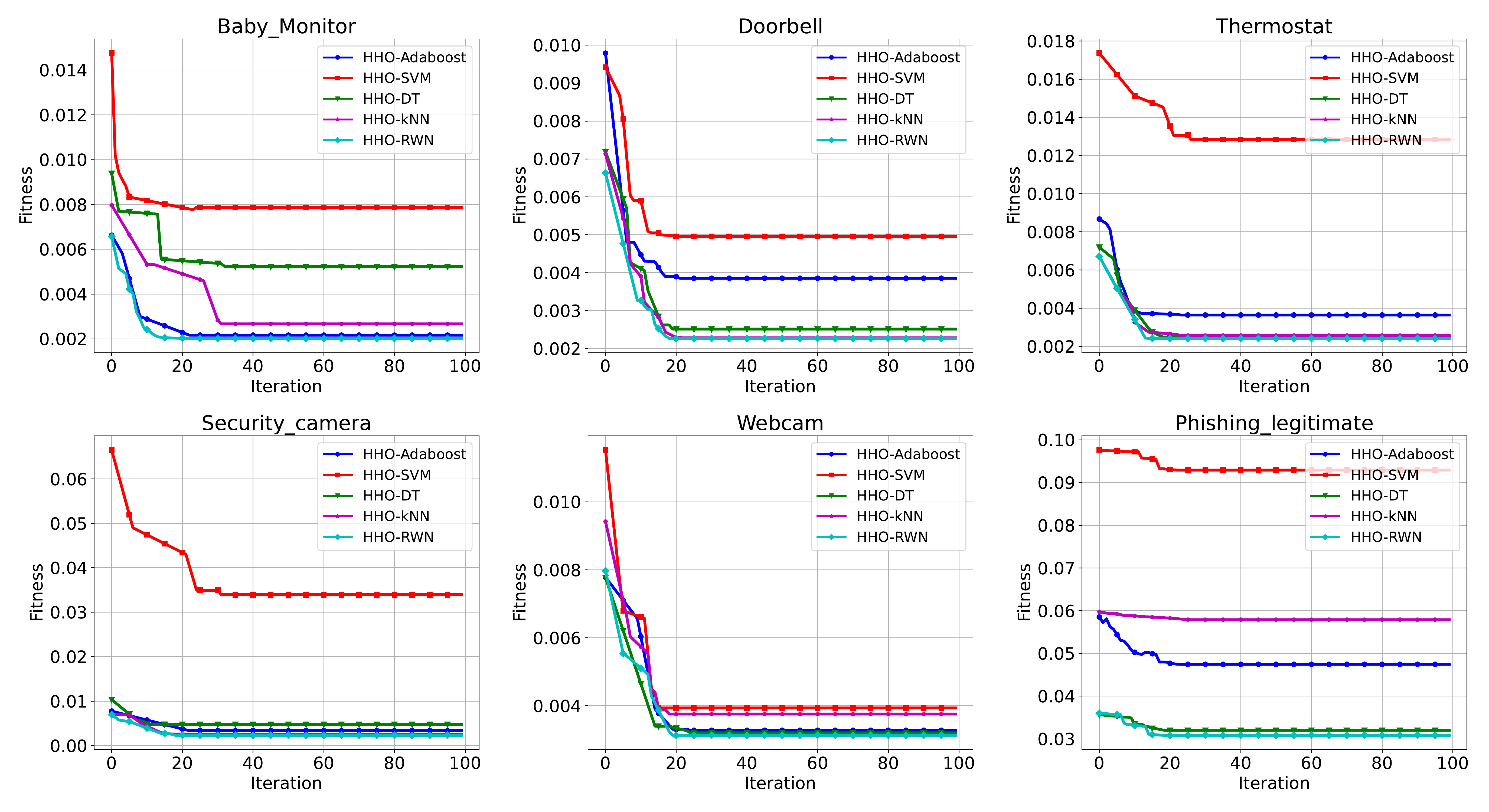}
\caption{The convergence curves for centralized ML models during training iterations.}
\label{fig:Conv_BM}       % Give a unique label
\end{figure} 
The maximum result accomplished by HHO-RWN with $99.9\%$. The F-measure rate has improved by $8.0\%$ compared to HHO-SVM in Webcam device.
Table \ref{tab:comp2} illustrates the results for the distributed ML model. We can observe that HHO-RWN and HHO-DT obtained the highest F-measure with $99.9\%$ in the Webcam dataset, and HHO-Adaboost achieved the second highest with $99.5\%$. The F-measure rate has improved up to $9.3\%$.
The precision rate has enhanced up to $14.5\%$ in Phishing legitimate dataset. According to the average recall results, the maximum result achieved by HHO-RWN with $99.0\%$, and the improvement rate reached $24.0\%$ in the Security camera device.
Additionally, we compare the performance of centralized and distributed ML models. 
The average F-measure of distributed ML yields the best results in the Security camera device with $99.9\%$, and it shares the same results with the centralized ML model in the Baby Monitor device with $99.8\%$ and in the Webcam device with $99.9\%$, and is very competitive in the other cases.
Overall, the results in the distributed ML scheme showed promising performance. In the distributed ML scheme, we trained the proposed approach on a portion of the training dataset. While in the centralized scheme, the proposed approach has been trained in the entire training dataset, leading to higher generalization and better performance.

\begin{figure}[t]
\centering
\includegraphics[width=0.50\textwidth ]{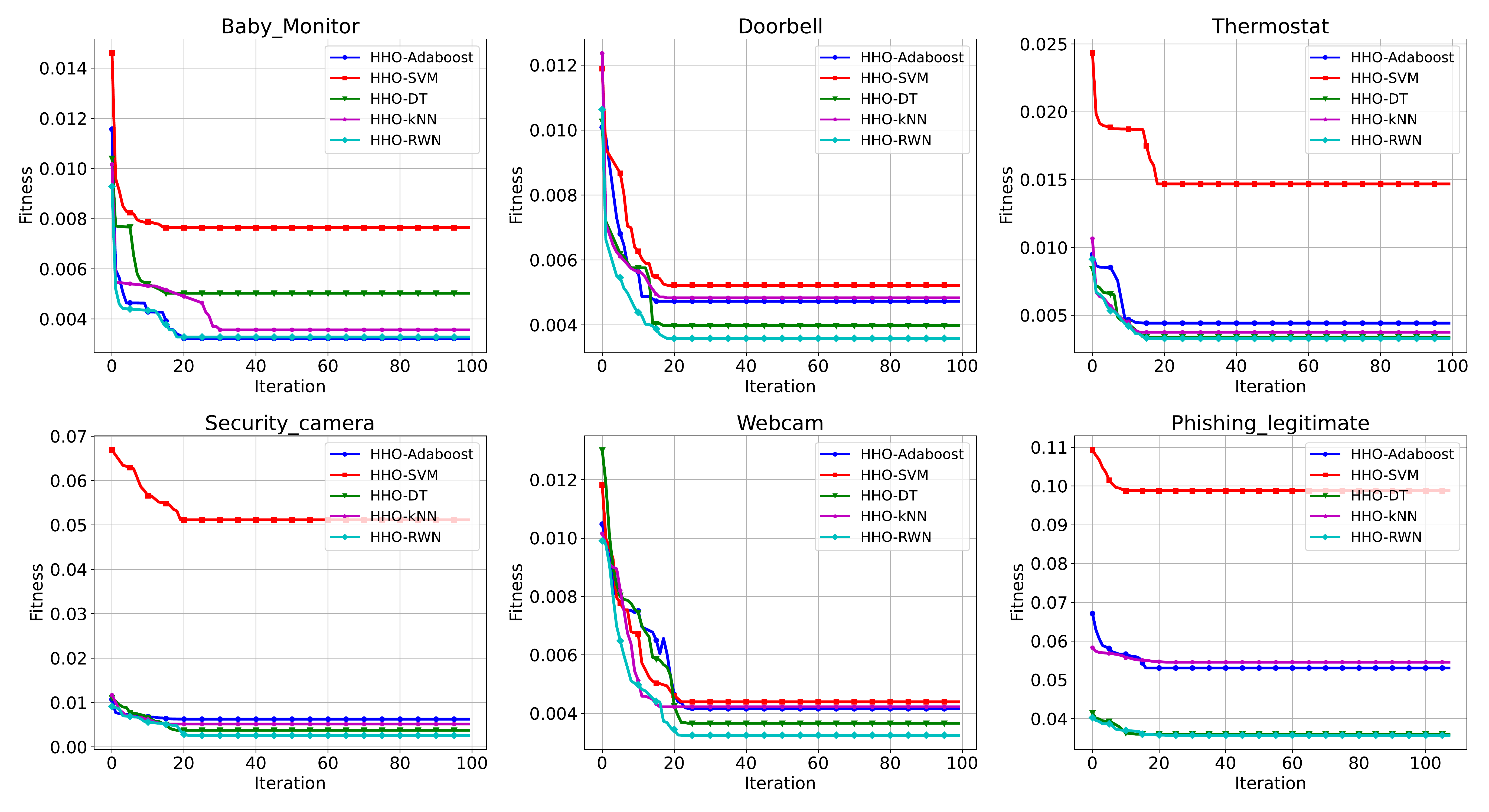}
\caption{The convergence curves for distributed ML models during training iterations.}
\label{fig:Conv_D}  
\end{figure}

The convergence curves of the HHO-RWN in centralized and distributed ML models with other classifiers are shown in Figs. \ref{fig:Conv_BM} and \ref{fig:Conv_D}, respectively for all datasets, according to the measurement of fitness. It can be seen that the different selection schemes show a similar convergence pattern. However, the proposed approach is relatively preferable with regards to the convergence rate. Noting that 100 iterations are used to monitor convergence in all cases.
It is reasonable to say that the best feature subsets and the best number of hidden neurons that are selected by the proposed approach enable it to obtain higher values based on the four metrics measured on the testing dataset.
Table \ref{tab:comparison} shows the comparison of average accuracy, recall, and precision values between HHO-RWN and other metaheuristics. According to the collected results, HHO-RWN achieved the highest average accuracy among all algorithms for two devices. HHO-RWN recorded the best performance with a 0.999\% accuracy at the Baby monitor device. When evaluating HHO-RWN ability to correctly identify anomalies, the results were the best in two devices. Table \ref{tab:comparison} also reveals that HHO-RWN achieved the best average precision values among two devices.

\begin{table*}[t]
\huge
\centering
\caption{Evaluation results of proposed approach and other comparative methods under the centralized learning scheme.}
\label{tab:comp}
\resizebox{\textwidth}{!}{%
\begin{tabular}{lllllllllllllllllllll}

\hline
                                      & \multicolumn{4}{c}{HHO-KNN}                                          & \multicolumn{4}{c}{HHO-SVM}                            & \multicolumn{4}{c}{HHO-Adaboost}                       & \multicolumn{4}{c}{HHO-DT}                                               & \multicolumn{4}{c}{HHO-RWN}                                       \\ \hline
\multicolumn{1}{l|}{Datasets}         & Acc            & Rec   & Prec  & \multicolumn{1}{l|}{F-measure}      & Acc   & Rec   & Prec  & \multicolumn{1}{l|}{F-measure} & Acc   & Rec   & Prec  & \multicolumn{1}{l|}{F-measure} & Acc            & Rec            & Prec  & \multicolumn{1}{l|}{F-measure} & Acc            & Rec            & Prec           & F-measure      \\ \hline
\multicolumn{1}{l|}{Baby\_Monitor}    & 0.996          & 0.989 & 0.993 & \multicolumn{1}{l|}{0.994}          & 0.991 & 0.954 & 0.991 & \multicolumn{1}{l|}{0.979}     & 0.997 & 0.992 & 0.988 & \multicolumn{1}{l|}{0.992}     & 0.990          & 0.993          & 0.993 & \multicolumn{1}{l|}{0.985}     & \textbf{0.999} & \textbf{0.996} & \textbf{0.995} & \textbf{0.998} \\
\multicolumn{1}{l|}{Doorbell}         & \textbf{0.999} & 0.989 & 0.987 & \multicolumn{1}{l|}{0.993} & 0.996 & 0.982 & 0.982 & \multicolumn{1}{l|}{0.954}     & 0.995 & 0.990  & 0.973 & \multicolumn{1}{l|}{0.992}     & 0.994          & \textbf{0.995} & 0.992 & \multicolumn{1}{l|}{0.977}     & 0.998          & 0.984          & \textbf{0.996} & \textbf{0.997} \\
\multicolumn{1}{l|}{Thermostat} & \textbf{0.999} & 0.958 & 0.978 & \multicolumn{1}{l|}{0.992}          & 0.984 & 0.983 & 0.984 & \multicolumn{1}{l|}{0.988}     & 0.998 & 0.851 & 0.972 & \multicolumn{1}{l|}{0.996}     & 0.998          & \textbf{0.995} & 0.933 & \multicolumn{1}{l|}{0.976}     & 0.995          & 0.968          & \textbf{0.995} & \textbf{0.999} \\
\multicolumn{1}{l|}{Security\_camera}       & 0.998          & 0.991 & 0.985 & \multicolumn{1}{l|}{0.970}          & 0.974 & 0.720 & 0.974 & \multicolumn{1}{l|}{0.963}     & 0.997 & 0.980 & 0.991 & \multicolumn{1}{l|}{\textbf{0.992}}     & 0.997          & \textbf{0.993} & 0.991 & \multicolumn{1}{l|}{0.960}     & \textbf{1.000}  & 0.992          & \textbf{1.000}  & 0.990 \\
\multicolumn{1}{l|}{Webcam}           & 0.997          & 0.992 & 0.989 & \multicolumn{1}{l|}{0.993}          & 0.996 & 0.978 & 0.996 & \multicolumn{1}{l|}{0.919}     & 0.998 & 0.992 & 0.982 & \multicolumn{1}{l|}{0.995}     & \textbf{0.999} & 0.993 & 0.992 & \multicolumn{1}{l|}{0.953}     & 0.998          & \textbf{0.997} & \textbf{0.998} & \textbf{0.999} \\ 
\multicolumn{1}{l|}{Phishing legitimate}           & 0.951          & 0.953 & 0.947 & \multicolumn{1}{l|}{0.951}          & 0.906 & 0.899 & 0.909 & \multicolumn{1}{l|}{0.821}     & 0.967 & 0.942 & 0.944 & \multicolumn{1}{l|}{0.944}    & 0.965 & 0.957 & 0.972 & \multicolumn{1}{l|}{0.965}     &\textbf{0.974}          & \textbf{0.973} & \textbf{0.975} & \textbf{0.974} \\ \hline
\end{tabular}%
}
\end{table*}
 
\begin{table*}[h]
\huge
\centering
\caption{Evaluation results of proposed approach and other comparative methods under the distributed learning scheme.}
\label{tab:comp2}
\resizebox{\textwidth}{!}{%
\begin{tabular}{lllllllllllllllllllll}

\hline
                                      & \multicolumn{4}{c}{HHO-KNN}                                          & \multicolumn{4}{c}{HHO-SVM}                            & \multicolumn{4}{c}{HHO-Adaboost}                       & \multicolumn{4}{c}{HHO-DT}                                               & \multicolumn{4}{c}{HHO-RWN}                                       \\ \hline
\multicolumn{1}{l|}{Datasets}         & Acc            & Rec   & Prec  & \multicolumn{1}{l|}{F-measure}      & Acc   & Rec   & Prec  & \multicolumn{1}{l|}{F-measure} & Acc   & Rec   & Prec  & \multicolumn{1}{l|}{F-measure} & Acc            & Rec            & Prec  & \multicolumn{1}{l|}{F-measure} & Acc            & Rec            & Prec           & F-measure      \\ \hline
\multicolumn{1}{l|}{Baby\_Monitor}    & 0.991          & \textbf{0.997} & 0.991 & \multicolumn{1}{l|}{0.996}          & 0.989 & 0.972 & 0.990 & \multicolumn{1}{l|}{0.994}     & 0.992 & 0.995 & \textbf{0.998} & \multicolumn{1}{l|}{0.996}     & 0.996          & 0.990          & 0.966 & \multicolumn{1}{l|}{0.990}     & \textbf{0.998} & 0.994 & 0.991 & \textbf{0.998} \\
\multicolumn{1}{l|}{Doorbell}         & \textbf{0.998} & \textbf{0.991} & 0.982 & \multicolumn{1}{l|}{0.997} & 0.995 & 0.989 & 0.982 & \multicolumn{1}{l|}{0.997}     & 0.993 & 0.984  & \textbf{0.994} & \multicolumn{1}{l|}{0.998}     & 0.993          & 0.990 & 0.970 & \multicolumn{1}{l|}{\textbf{0.999}}     & 0.996          & 0.980          & \textbf{0.994} & 0.996 \\
\multicolumn{1}{l|}{Thermostat} & \textbf{0.997} & 0.945 & 0.926 & \multicolumn{1}{l|}{0.997} & 0.984 & 0.980 & 0.984 & \multicolumn{1}{l|}{0.981}  & 0.996 & 0.943 & 0.965 & \multicolumn{1}{l|}{0.997}     & \textbf{0.997} & 0.895 & 0.969 & \multicolumn{1}{l|}{\textbf{0.998}} & 0.992 & \textbf{0.971}  & \textbf{0.993} & 0.997 \\
\multicolumn{1}{l|}{Security\_camera}       & 0.998          & 0.988 & 0.988 & \multicolumn{1}{l|}{0.971}          & 0.970 & 0.750 & 0.950 & \multicolumn{1}{l|}{0.924}     & 0.997 & 0.983 & 0.992 & \multicolumn{1}{l|}{0.994}     & \textbf{0.999} & 0.981 & 0.983 & \multicolumn{1}{l|}{\textbf{0.998}}     & 0.997  & \textbf{0.990} & \textbf{0.997}  & \textbf{0.998} \\
\multicolumn{1}{l|}{Webcam}           & 0.996          & 0.990 & 0.992 & \multicolumn{1}{l|}{0.990} & 0.990 & 0.988 & 0.994 & \multicolumn{1}{l|}{0.906}     & 0.996 & 0.996 & \textbf{0.996} & \multicolumn{1}{l|}{0.995} & \textbf{0.999} & \textbf{0.998} & \textbf{0.996} & \multicolumn{1}{l|}{\textbf{0.999}}     & 0.995          & 0.996 & 0.994 & \textbf{0.999} \\ 
\multicolumn{1}{l|}{Phishing legitimate}           & 0.941          & 0.886 & 0.927 & \multicolumn{1}{l|}{0.930}          & 0.909 & 0.857 & 0.827 & \multicolumn{1}{l|}{0.898}  & 0.969 & 0.958 & 0.959 & \multicolumn{1}{l|}{0.958}    & 0.959 & 0.936 & 0.935 & \multicolumn{1}{l|}{0.959}     &\textbf{0.969}          & \textbf{0.970} & \textbf{0.972} & \textbf{0.963} \\ \hline
\end{tabular}%
}
\end{table*}

\begin{table*}[t]
\huge
\caption{Comparison between HHO-RWN and other metaheuristics in terms of Accuracy, Recall, and Precision}
\label{tab:comparison}
\resizebox{\textwidth}{!}{%
\begin{tabular}{lllllllllllll}
\hline
\multicolumn{1}{c}{}                     & \multicolumn{4}{c}{Accuracy}                                            & \multicolumn{4}{c}{Recall}                                                                   & \multicolumn{4}{c}{Precision}                          \\ \hline
\multicolumn{1}{c|}{Dataset}             & HHO-RWN        & NSGA-II & MOPSO-Lévy & \multicolumn{1}{l|}{SPEA-II}    & HHO-RWN        & NSGA-II    & MOPSO-Lévy     &  \multicolumn{1}{l|}{SPEA-II} & HHO-RWN        & NSGA-II & MOPSO-Lévy     & SPEA-II    \\ \hline
\multicolumn{1}{l|}{Baby Monitor}        & \textbf{0.999} & 0.565   & 0.937      & \multicolumn{1}{l|}{0.604}      & \textbf{0.996}          & 0.985      & 0.920          & \multicolumn{1}{l|}{0.967}   & \textbf{0.995}          & 0.479   & 0.710          & 0.521      \\
\multicolumn{1}{l|}{Doorbell}            & \textbf{0.998} & 0.487   & 0.989      & \multicolumn{1}{l|}{0.801}      & \textbf{0.984}          & 0.693      & 0.780          & \multicolumn{1}{l|}{0.902}    & 0.996          & 0.256   & \textbf{1.000} & 0.559      \\
\multicolumn{1}{l|}{Thermostat}          & 0.995 & 0.289   & \textbf{0.998}      & \multicolumn{1}{l|}{0.327}      & 0.968          & 0.149      & \textbf{0.997} &  \multicolumn{1}{l|}{0.157}   & \textbf{0.995} & 0.022   & 0.957          & 0.000           \\ \hline
\end{tabular}%
}
\end{table*}

Since privacy and security factors are essential in many applications, distributed ML is a promising learning solution. It enables the training of machine learning models locally, so there is no need to transfer the data to a central cloud. Thus, the reduction in heavy data transmissions will minimize the number of potential attacks consequently decreasing the risks of privacy leakage.
Therefore, the choice of whether to build a centralized or distributed model depends on the target application and the nature of the data. We can say that choosing the appropriate learning scheme depends on the application's environment, priorities, and objectives while considering factors of privacy, performance, and computational resources.

\section{Conclusion}\label{sec:conclusion}
Due to the dramatic increase in the number of IoT devices deployed online and the security sensitives of these devices, hackers and attackers exploit the increase of such devices to deploy different types of attacks. Intrusion detection systems are one of the techniques allocated to protecting the IoT network. The main aim of the proposed approach is to detect IoT botnet attacks by utilizing HHO to improve RWN and at the same time, to perform FS based on centralized and distributed ML models. Experimental results revealed that the proposed model performs well in centralized and distributed models to detect attacks on smart IoT devices.
For future work, we plan to integrate the proposed approach using Federated Learning (FL) to improve security and privacy issues. In addition, investigate the efficiency of the proposed approach for other types of applications such as the healthcare system.

\section*{Acknowledgements}
This research was supported by the Technology Innovation
Institute (TII), Abu Dhabi, UAE, under the CyberAI project (grant
number: TII/DSRC/2022/3036).

\balance
\bibliographystyle{IEEEtran}
\bibliography{Ref}
\end{document}